\documentclass[11pt]{article}
\usepackage[T1]{fontenc}
\usepackage{fancyhdr,graphicx,times,microtype,geometry,amsmath}
\usepackage{amsfonts}
\usepackage{amssymb}
\usepackage{amsmath,bm}

\usepackage{mathrsfs}

\usepackage{scalerel}

\newcommand\scale[2]{\vstretch{#1}{\hstretch{#1}{#2}}}
	
\newcommand{\LIPplus}{\if@draft
		\mathbin{\ooalign{$\bigtriangleup$\crcr\hidewidth \raise.14em\hbox{$\scale{0.7}{\scriptscriptstyle+}$}\hidewidth}}
	\else
		\mathbin{\mathpalette\LIPcls+}
	\fi}
\newcommand{\LIPminus}{\if@draft
		\mathbin{\ooalign{$\bigtriangleup$\crcr\hidewidth \raise.14em\hbox{$\scale{0.7}{\scriptscriptstyle-}$}\hidewidth}}
	\else
		\mathbin{\mathpalette\LIPcls-}
	\fi}
\newcommand{\LIPtimes}{\if@draft
	  \mathbin{\ooalign{$\bigtriangleup$\crcr\hidewidth \raise.14em\hbox{$\scale{0.7}{\scriptscriptstyle\times}$}\hidewidth}}
	\else
		\mathbin{\mathpalette\LIPcls\times}
	\fi}

\newcommand{\LIPcls}[2]{%
  \ooalign{$#1\bigtriangleup$\crcr  \hidewidth\raisefix{#1}\hbox{$#1\scale{0.45}{\bm{#2}}$}\hidewidth}}

\def\raisefix#1{%
  \ifx#1\displaystyle
    \raise.14em
  \else
    \ifx#1\textstyle
      \raise.14em
    \else
      \ifx#1\scriptstyle
        \raise.112em
      \else
        \raise.0933em
      \fi
    \fi
  \fi
}

\newcommand{\LP}{\LIPplus}
\newcommand{\LM}{\LIPminus}
\newcommand{\LT}{\LIPtimes}

\newcommand{\Real}{\mathbb R}

\newcommand{\la}{\lambda}

\DeclareGraphicsExtensions{.pdf,.png,.jpg,.jpeg}

\begin{document}

\begin{center}
{\large{\bf \MakeUppercase{HOMOGENEITY OF A REGION IN THE LOGARITHMIC IMAGE PROCESSING FRAMEWORK: APPLICATION TO REGION GROWING ALGORITHMS} }}\\[1ex]
Michel Jourlin {\scriptsize $^{\textrm{a,c}}$}, Guillaume Noyel {\scriptsize $^{\textrm{b,c}}$}\\
{\small {\scriptsize $^{\textrm{a}}$} Lab. H. Curien, UMR CNRS 5516, 18 rue Pr. B. Lauras, 42000 St-Etienne, France, michel.jourlin@univ-st-etienne.fr;\\ 
{\scriptsize $^{\textrm{b}}$} University of Strathclyde Institute of Global Public Health, Lyon Ouest Ecully, France, guillaume.noyel@i-pri.org;\\
{\scriptsize $^{\textrm{c}}$} International Prevention Research Institute, iPRI, Lyon, France}
\end{center}

\bigskip

{\large \textbf{Abstract:} The current paper deals with the role played by Logarithmic Image Processing (LIP) operators for evaluating the homogeneity of a region. Two new criteria of heterogeneity are introduced, one based on the LIP addition and the other based on the LIP scalar multiplication. Such tools are able to manage Region Growing algorithms following the Revol's technique \cite{Revol1997}: starting from an initial seed, they consist of applying specific dilations to the growing region while its inhomogeneity level does not exceed a certain level. The new approaches we introduce are significantly improving Revol's existing technique by making it robust to contrast variations in images. Such a property strongly reduces the chaining effect arising in region growing processes.}\\[1ex]

\textbf{Keywords:} Logarithmic Image Processing, LIP, homogeneity, heterogeneity criterion, region growing, chaining effect, segmentation, low-contrasted images. 

%
%
\section{Introduction, recalls and notations}

If $f$ and $g$ represent two grey-level functions defined on $D \subset \Real^2$ with value in the grey scale $[0, M[$, $M \in \Real$ and $\la$ a real number, let us remember that the LIP operators \cite{Jourlin2016} are defined by the formulas:
\begin{itemize}
	\item addition of $f$ and $g$:
	\begin{equation}
		f \LP g = f + g - f.g/M. \label{eq:LIP:plus}%
	\end{equation}
	\item subtraction of $f$ by $g$:
	\begin{equation}
		f \LM g = (f-g)(1-g/M). \label{eq:LIP:minus}
	\end{equation}
	\item scalar multiplication of $f$ by $\la$:	
	\begin{equation}
		\lambda \LIPtimes f = M - M \left( 1 - f/M \right)^{\lambda}. \label{eq:LIP:times}%
	\end{equation}
\end{itemize}
For each of these laws, a notion of \textit{Logarithmic Contrast} has been defined \cite{Jourlin2012}:
\begin{itemize}
	\item the Logarithmic Additive Contrast (LAC):
	\begin{equation}
		C^{\LP}_{(x,y)}(f) = LAC_{(x,y)}(f) = \sup{(f(x),f(y))} \LM \inf{(f(x),f(y))} = \frac{|f(x)-f(y)|}{1-\frac{\inf{(f(x),f(y))}}{M}}. \label{eq:LIP:LAC}%
	\end{equation}
	\item the Logarithmic Multiplicative  Contrast (LMC):
	\begin{equation}
		C^{\LT}_{(x,y)}(f) = LMC_{(x,y)}(f) = \frac{\ln{\left(1-\frac{\sup{(f(x),f(y))}}{M}\right)}}{\ln{\left(1-\frac{\inf{(f(x),f(y))}}{M}\right)}}. \label{eq:LIP:LMC}%
	\end{equation}	
\end{itemize}

%
%
\section{Heterogeneity of a region in the LIP framework}  

Given a grey level function $f$ and a region $R$ of its domain $D$, let us define:
\begin{itemize}
	\item the LIP-additive Heterogeneity of $R$
	\begin{equation}
		H^{\LP}_{f}(R) = LAC(\sup_{x \in R}{f(x)} , \inf_{x \in R}{f(x)}) = \sup_{x \in R}{f(x)} \LM \inf_{x \in R}{f(x)}, \label{eq:LIP-additive_heterogeneity}%
	\end{equation}
	\item the LIP-multiplicative Heterogeneity of $R$
	\begin{equation}
		H^{\LT}_{f}(R) = LMC(\sup_{x \in R}{f(x)} , \inf_{x \in R}{f(x)})  =  \frac{\ln{\left(1-\frac{\sup_{x \in R}{f(x)}}{M}\right)}}{\ln{\left(1-\frac{\inf_{x \in R}{f(x)}}{M}\right)}}. \label{eq:LIP-multiplicative_heterogeneity}%
	\end{equation}
\end{itemize}

%
%
\section{Application to Region Growing}

Let $R_{n}$ be a region built after $n$ iterations: according to Revol's approach \cite{Revol1997,Jourlin2016}, all the neighbouring pixels of $R_{n}$ are in a first stage aggregated to it, resulting in the dilation $R_{n} \oplus N$ of $R_{n}$ by the considered neighbourhood $N$ (generally $N_8$ constituted of the 8 nearest pixels in a square grid) \cite{Serra1982}. The question arising then is to determine if $R_{n} \oplus N$ is considered homogeneous or not.
Therefore, we compute one of the heterogeneity parameters previously defined, for example $H_f^{\LP} (R_{n} \oplus N)$:
\begin{itemize}
	\item if $H_f^{\LP} (R_{n} \oplus N) \leq t$ where $t$ is a given threshold, the new region becomes $R_{n+1}= R_{n} \oplus N$,
	\item if $H_f^{\LP} (R_{n} \oplus N) > t$ the strategy consists of removing from $R_{n} \oplus N$ the most penalizing points, namely the points $y$ which satisfy $f(y)=\sup_{x \in R_{n} \oplus N} f(x)$ or $f(y)=\inf_{x \in R_{n} \oplus N} f(x)$ until the homogeneity criterion becomes true again, producing the new region $R_{n+1}$.
\end{itemize}
The technique is applied until the obtained region does not grow any more (i.e. is the same than the previous one).

%
%
\section{Illustration}

\begin{figure}[h]
\begin{center}
\begin{tabular}{@{}c@{ }c@{ }c@{}}
\includegraphics[width=0.33\columnwidth]{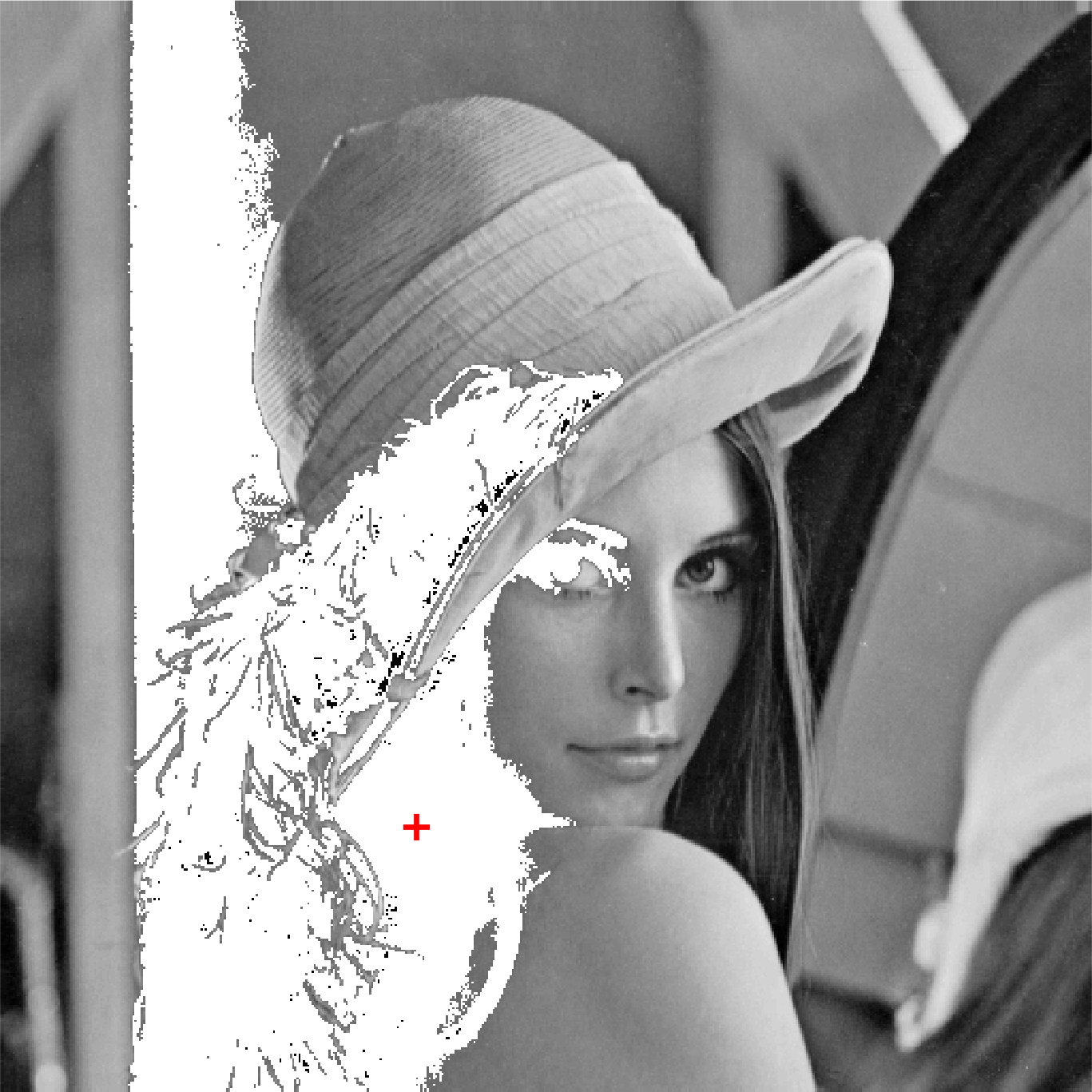}&
\includegraphics[width=0.33\columnwidth]{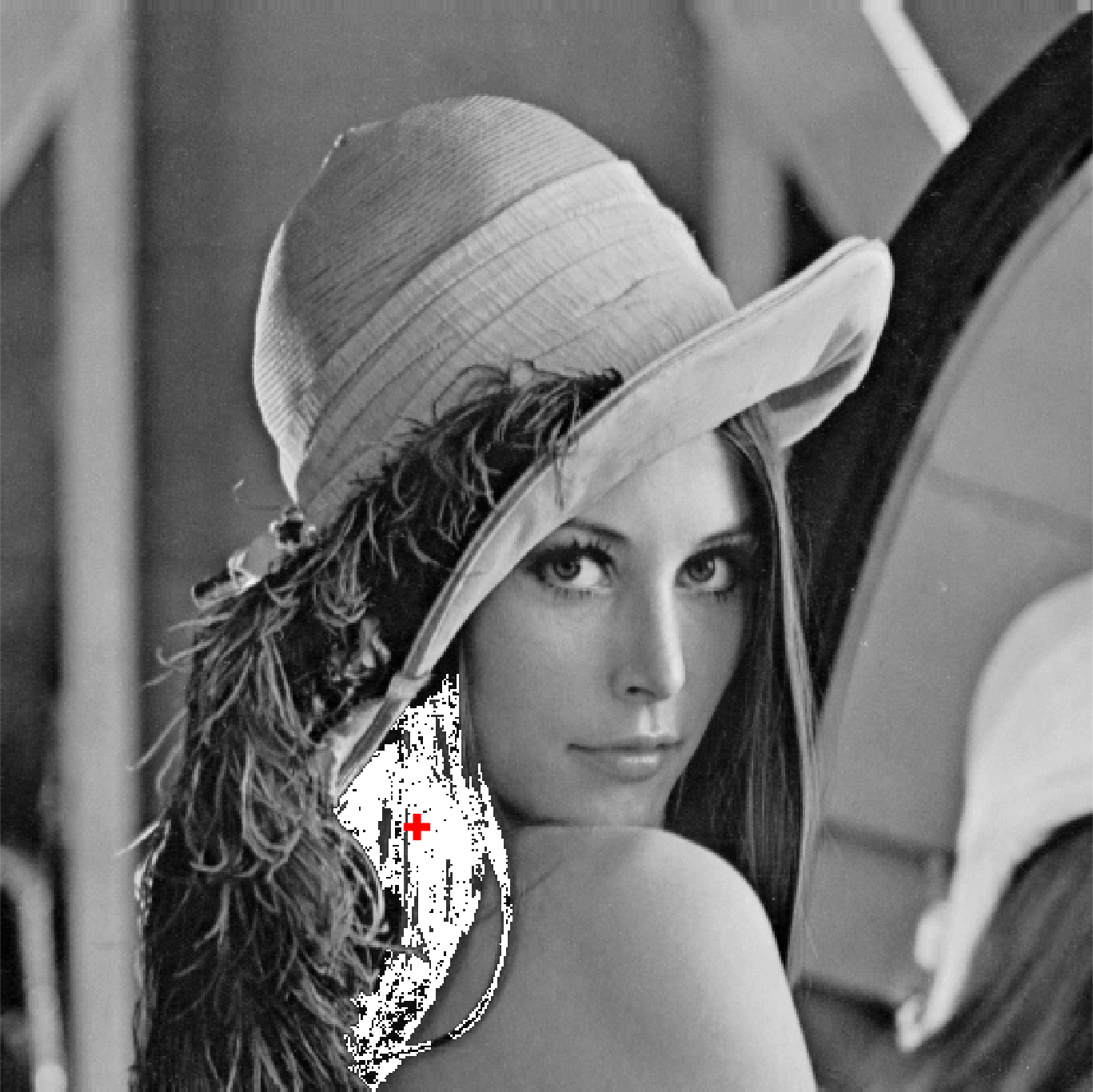}&
\includegraphics[width=0.33\columnwidth]{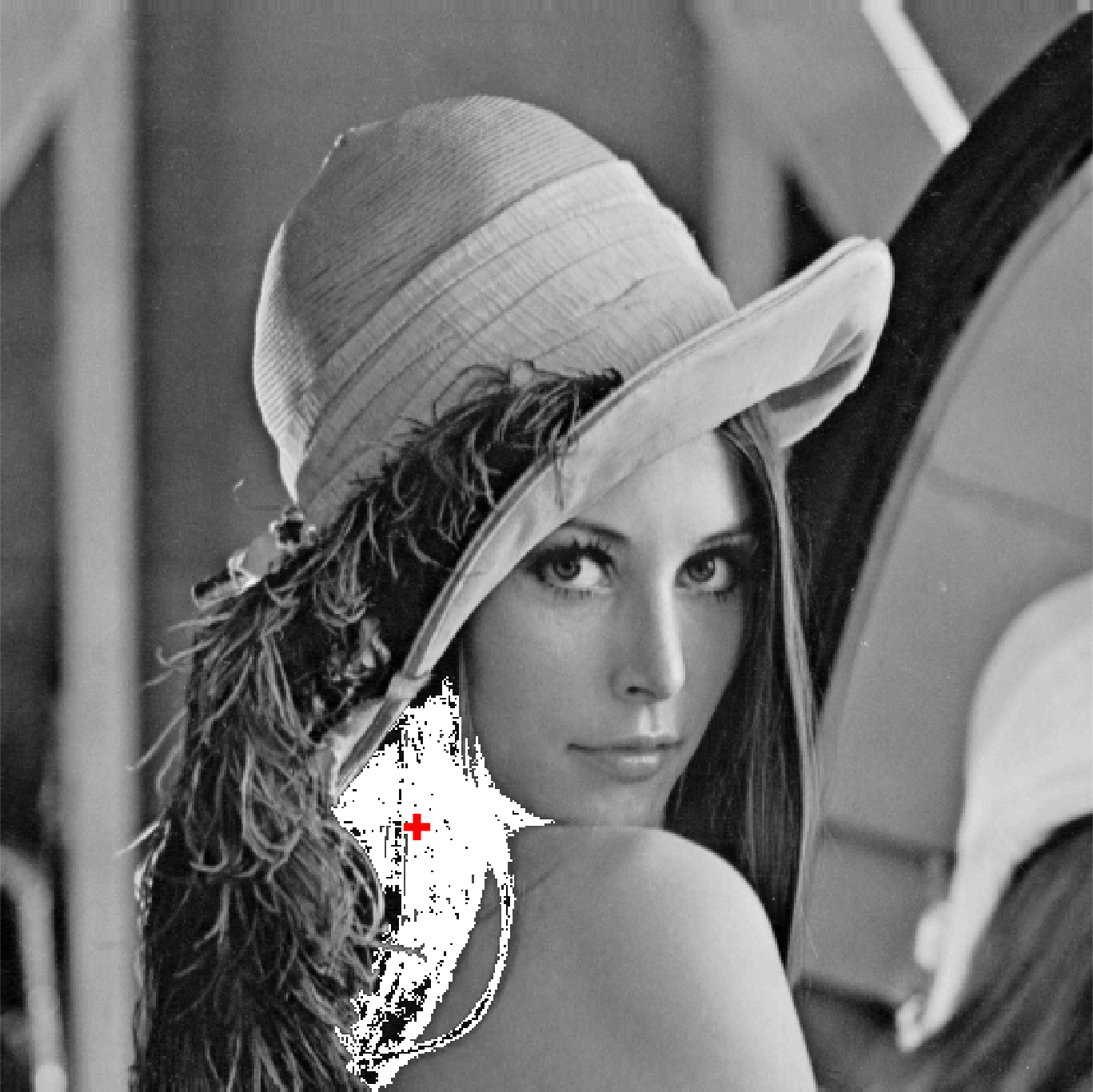}\\
(a) & (b) & (c)\\
\end{tabular}
\caption{Segmentation by region growing. The seed point is located at the red cross. (a) Revol's method (t =25). (b) Our approach with the LIP-additive heterogeneity criterion $H^{\protect \LP}_{f}$ (t =25). (c) Our approach with the LIP-multiplicative heterogeneity criterion $H^{\protect \LT}_{f}$ (t =2).}
\label{fig:RG_lena}
\end{center}
\end{figure}

In Figure \ref{fig:RG_lena}, a seed point is selected in the hair part of the ``Lena'' image in order to be the starting point of the growing process algorithm. When using a threshold $t=25$, in Revol's approach (Fig. \ref{fig:RG_lena} a), the hair area is aggregated with other parts such as the (feather boa) scarf or a part of the face. This is the \textit{chaining effect}. With our approach, when using the LIP-additive heterogeneity criterion $H^{\protect \LP}_{f}$ (Fig. \ref{fig:RG_lena} b), only the hair area is aggregated. When using the LIP-multiplicative heterogeneity criterion $H^{\protect \LT}_{f}$ (Fig. \ref{fig:RG_lena} c), with $t=2$, only the hair area is aggregated. Therefore, our approach limits the chaining effect that can occur in Revol's approach especially when there are small contrast variations in between regions.

%
%
\section{Conclusion and perspectives}

Two new heterogeneity criteria of a region are introduced, namely the LIP-additive heterogeneity and the LIP-multiplicative heterogeneity criterion. Both are tested in ``Lena'' image and compared to an existing method: Revol's approach. Results show that the new criteria are more robust to chaining effect especially in low-contrasted region. Indeed, the LIP model used for these criteria possesses the property of dealing with low-contrasted region in a similar way to normally contrasted region. Such heterogeneity criteria strengthen the region growing process of Revol's and open the way to numerous applications where the illumination is not or partially controlled.

\bibliographystyle{abbrv}
\bibliography{refs}

\end{document}